\documentclass[a4paper]{article}
\usepackage[final]{pdfpages}
\usepackage{INTERSPEECH_v2,multirow,array,xcolor,pgfplots,balance,setspace,amsmath,graphicx,multirow}
\usepackage{tikz}
\usetikzlibrary{positioning, fit, arrows.meta,shapes.geometric, arrows, shapes, decorations.text,calc}

\newcolumntype{C}[1]{>{\centering\let\newline\\\arraybackslash\hspace{0pt}}m{#1}}

\title{The CAPIO 2017 Conversational Speech Recognition System}

\name{Kyu J. Han, Akshay Chandrashekaran, Jungsuk Kim, Ian Lane}
\address{
  Capio Inc., Belmont, CA, USA}
\email{\{kyu,akshay,jungsuk,ian\}@capio.ai}

\begin{document}
\maketitle
\begin{abstract}
In this paper we show how we have achieved the state-of-the-art performance on the industry-standard NIST 2000 Hub5 English evaluation set. We propose densely connected LSTMs, (namely, dense LSTMs), inspired by the densely connected convolutional networks recently introduced for image classification tasks. It is shown that the proposed dense LSTMs would provide more reliable performances as compared to the conventional, residual LSTMs as more LSTM layers are stacked in neural networks. We also propose an acoustic model adaptation scheme that simply averages the parameters of a seed neural network acoustic model and its adapted version. This method was applied with the CallHome training corpus and improved individual system performances by on average 6.1\% (relative) against the CallHome portion of the evaluation set with no performance loss on the Switchboard portion. With RNN-LM rescoring and lattice combination on the 5 systems trained across three different phone sets, our 2017 speech recognition system has obtained 5.0\% and 9.1\% on Switchboard and CallHome, respectively, both of which are the best word error rates reported thus far. According to IBM in their latest work to compare human and machine transcriptions, our reported Switchboard word error rate can be considered to surpass the human parity (5.1\%) of transcribing conversational telephone speech. We also share the performance numbers of our system on non-telephony environments for readers' benefits.
\end{abstract}

\section{Introduction}

We have recently observed a series of leap-frog advancements in deep learning based acoustic and language modeling for conversational speech recognition. With the contributions mainly from convolutional neural networks (CNNs) and recurrent neural networks (RNNs), multiple research groups have continued to improve their system performance on the well-known, industry-standard NIST 2000 Hub5 English evaluation set\footnote{As known as Switchboard, but it actually consists of the two testsets of Switchboard and CallHome.} \cite{Xiong16,Saon17,Han17,Xiong17,Kurata17}, approaching to the hypothesized human performance of the evaluation set. Achieving human parity has now become the topic of the speech recognition community, which nurtured interesting research works of contrasting transcriptions from human transcribers and conversational speech recognition systems \cite{Xiong16,Saon17,Stolcke17}. It is reported in \cite{Stolcke17} that similar error patterns were found between human and machine transcriptions, hinting that the quality of machine transcriptions becomes closer to that of human transcribers.

Conversational speech recognition, however, is still challenging, and we in \cite{Han17} performed a comparative analysis on how vulnerable even the state-of-the-art conversational speech recognition system would be against real-world telephone conversations in the wild. One of the causes to make it hard for speech recognition systems to perform well enough against real-world data is the acoustic and lexical variability of the real-world data that are not exposed in a training phase. To overcome this mismatch, neural network acoustic models need to be adapted, but it is widely known that they are not easy to be adapted due to a large number of parameters to be tuned. Most of the research effort on neural network adaptation thus has been focused either to update a part of parameters while fixing the rest \cite{Li10,Liao13,Himawan15,Lu16} or to append domain-specific features (e.g., i-vectors \cite{Dehak11} in case of speaker adaptation) for a better feature transformation \cite{Saon13,Gupta14,Miao14}. In this paper, we propose a simple but generalized adaptation method for deep neural networks such that it can obtain expected adaptation benefits as well as avoids overfitting. We applied this scheme with the CallHome training corpus to our individual systems and observed the performance improvement of on average 6.1\% (relative) on the CallHome subset of the NIST 2000 Hub5 English evaluation set without any loss on the Switchboard subset.

We also propose a new neural network based acoustic model structure with dense connections between long short-term memory (LSTM) layers. Densely connected neural networks were originally introduced to avoid layer-wise vanishing gradient problems when CNNs are stacked in a very deep fashion, e.g., more than 100-layers, for image recognition tasks \cite{Huang16}. One can view dense connection as a variant from residual learning \cite{He16} or highway networks \cite{Srivastava15a,Srivastava15b}. In speech recognition, residual or highway connection have been applied to LSTMs, only between adjacent layers \cite{Zhang16,Pundak17,Kim17,Huang17}. Our dense LSTMs connect (almost) every layer to one another to mitigate vanishing gradient effect between LSTM layers and help error signals propagated even back to the very first layer during back propagation in training. Benefiting from the proposed dense LSTMs, we were able to reach the marks of 5.0\% and 9.1\% in word error rate (WER) for the Switchboard and CallHome testsets, respectively, both of which are the best results reported thus far in the field.

This paper is organized as follows. Section 2 describes the proposed, densely connected LSTMs, accompanying empirical analysis on residual and dense LSTMs. It can give an insight of how dense connections in LSTM help to keep improving accuracy as more LSTM layers are stacked in neural networks for speech recognition tasks. Section 3 highlights the proposed acoustic model adaptation scheme, and shares the performance of individual systems before and after model adaptation using the CallHome training data. In Section 4, we detail the other components constituting our 2017 conversational speech recognition system, such as language models and system combination. We present experimental results in a broader scale across individual systems in Section 5, in a view point of the industry-standard NIST 2000 Hub5 English evaluation set, which is telephony data. In addition, we share the performances of our system on non-telephony environments in Section 6, focusing on the two well-known data sets in TED-LIUM and LibriSpeech. In Section 7, we conclude this paper with the remarks on the contributions and future directions.

\section{Densely Connected LSTM}
\label{sec:dense}

Dense connection \cite{Huang16} was introduced for CNNs to yield the state-of-the-art performance on the CIFAR-10/100 data sets \cite{Krizhevsky09} for image classification, outperforming residual networks \cite{He16,Zagoruyko16} which had been the best performing neural network architecture in the domain. Like skip connections in residual learning, dense connections let error signals further back-propagated with less gradient vanishing effect between layers in a deep neural network. One notable difference between dense networks and residual networks is a connectivity pattern. Considering that $H_{\ell}(\cdot)$ is a general composite function of operations in the $\ell^{\text{th}}$ layer of a given neural network, a residual connectivity for the output of the $\ell^{\text{th}}$ layer, $\mathbf{x}_{\ell}$, can be written as

\begin{figure}[t]
\begin{tikzpicture} \hspace{-0.1cm}
\pgfplotsset{width=7.5cm}
\begin{axis}[
    xlabel={Number of LSTM Layers},
    ylabel={Word Error Rate (\%)},
    xmin=1, xmax=30,
    ymin=15, ymax=30,
    xtick={1,5,10,15,20,25,30},
    ytick={15.0,16,17.5,20.0,25.0,30.0},
    legend pos=north west,
    legend style={nodes={scale=0.65, transform shape}},
    label style={font=\small},
    ymajorgrids=true,
    grid style=dashed
]
  \addplot[
    color=red,
    mark=o,
    ]
    coordinates {
    (1,24.8)(3,19.9)(5,19.5)(6,19.1)(7,19.3)(8,20.8)(9,23.1)(10,24.8)
    };
    \addlegendentry{LSTM}
 \addplot[
    color=orange,
    mark=o,
    ]
    coordinates {
    (1,24.8)(3,19.4)(5,18.5)(6,18.1)(7,17.9)(8,17.9)(9,17.9)(10,17.7)(15,18.2)(20,20.0)(30,29.8)
    };
    \addlegendentry{Residual LSTM}
  \addplot[
    color=teal,
    mark=o,
    ]
    coordinates {
    (1,24.8)(5,18.7)(10,17.5)(15,17.6)(20,17.7)(30,20.4)
    };
    \addlegendentry{ Dense LSTM (5 / block)}
  \addplot[
    color=cyan,
    mark=o,
    ]
    coordinates {
    (1,24.8)(10,16.8)(20,16.3)(30,16.4)
    };
    \addlegendentry{Dense LSTM (10 / block)}
\end{axis}
\end{tikzpicture}
\caption{WER comparison between residual and dense connection for LSTMs with the cell dimension of 128.}
\end{figure}
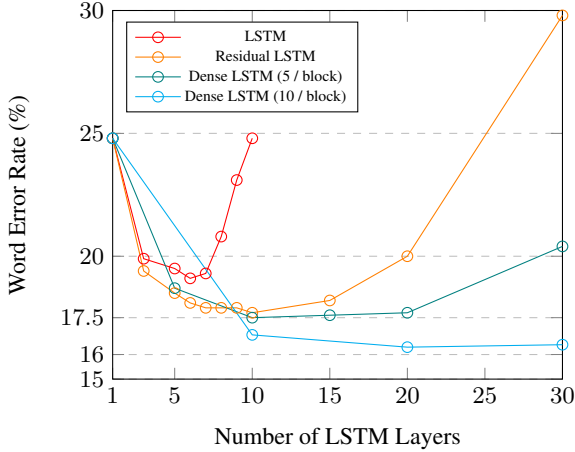

\begin{equation}
\mathbf{x}_{\ell} = H_{\ell} ( \mathbf{x}_{\ell-1} ) + \mathbf{x}_{\ell-1},
\end{equation}
while a dense connectivity can be represented as

\begin{equation}
\mathbf{x}_{\ell} = H_{\ell} ( \left[ \mathbf{x}_1 , \mathbf{x}_2 , \cdot \cdot \cdot \mathbf{x}_{\ell-1} \right] ),
\end{equation}
where $[ \mathbf{x}_1 , \mathbf{x}_2 , \cdot \cdot \cdot \mathbf{x}_{\ell-1} ] $ is a concatenated vector of outputs from the first layer to the $({\ell - 1})^{\text{th}}$ layer. The dense connectivity pattern accommodates more direct connections throughout layers while residual connections are only made between adjacent layers. 



We propose \textit{densely connected LSTMs} (namely, dense LSTMs) in this paper, inspired by the success of dense connection for CNNs. In speech recognition, there has been a limited number of effort to exploit residual connection or its variants, e.g., highway connection, to LSTMs with minor differences in implementation \cite{Zhang16,Pundak17,Kim17,Huang17}, but none using dense connection yet. To understand how dense LSTMs would work as layers get deeper, let us take a look at Figure 1. For the experiments, we trained (uni-directional) LSTMs with the cell dimension of 128 using a small portion of our entire training data, i.e., 300hr Switchboard-1 Release 2 corpus from LDC (LDC97S62), and tested them against the NIST 2000 Hub5 English evaluation set (Switchboard and CallHome combined). The red curve indicates that normal LSTMs would not obtain any benefit after the 6th layer where the lowest WER of 19.1\% is reached. The performance of residual LSTMs, depicted as the orange curve, seems further improved until the 10th layer where 17.7\% is marked and then continues to degrade thereafter as more layers are added. This validates that residual learning makes LSTMs perform better with more layers as has been reported in \cite{Zhang16,Pundak17,Kim17,Huang17}, but we also see that there is a clear limitation. The U-shaped curve might be the reason why most of the residual LSTMs for speech recognition so far have been stuck with only up to 15 layers with marginal improvement in WER. In contrast, dense LSTMs are shown continuously benefited as more layers are added even after the 10th layer, further pulling the lowest possible WER down to around 16\% at the number of LSTM layers of 20 (light blue curve). There are a couple of notes on the dense LSTMs experimented. Due to the connectivity pattern in Eq. (2) of concatenating vectors coming out of the previous layers, the dimension of an input vector for the $\ell^\text{th}$ LSTM layer with the cell dimension of $d$ is $(\ell-1) \times d$, which would keep increasing as $\ell$ goes larger. Thus we grouped LSTM layers into blocks where dense connections are applied only within the same block, and linked blocks by a transitional layer. This dense block concept was also exploited in the original paper for densely connected CNNs \cite{Huang16}, but with other purposes. The green curve in Figure 1 is based on the dense LSTMs where every group of 5 LSTM layers belong to one block while the light blue curve comes from the dense LSTMs with 10 LSTM layers per block.

\begin{figure}[t]
    \label{fig:denseTdnnBlstm}
    \begin{minipage}[b]{0.9\linewidth}
        \centering
        \includegraphics[width=8cm]{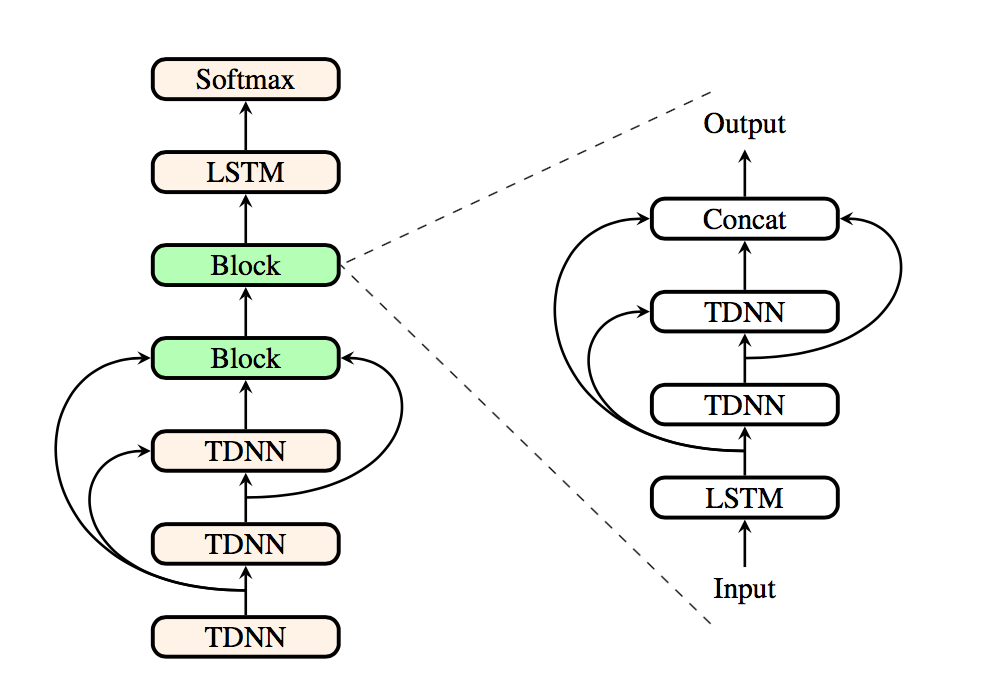}
    \end{minipage}
    \caption{Structure of a dense TDNN-LSTM acoustic model. Each dense block outputs 1,024 dimensional non-linear activation vectors.}
\end{figure}

\begin{table}
    \centering
    \caption{WER comparison for TDNN-LSTMs with and without dense connection.}
    \vspace{3mm}
    \begin{tabular}{c|c|c}
        \hline
         \textbf{Acoustic Model} & \textbf{SWBD} & \textbf{CH} \\
         \hline
         \hline
         \small TDNN-LSTM & \small 7.3\% & \small 13.8\%\\
         \hline
         \small Dense TDNN-LSTM & \small 7.3\% & \small 13.0\% \\
         \hline
    \end{tabular}
    \label{tab:dense}
\end{table}

Dense connection can be easily applied to existing LSTM-based neural network architectures for speech recognition, thanks to a simple connectivity pattern. It can improve performance as more LSTM layers are added, since it helps alleviate layer-wise vanishing gradient effect. Based on our experimental validation from Figure 1, we propose two dense LSTM architectures for conversational speech recognition, which are detailed in the next subsections.

\subsection{Dense TDNN-LSTM}
The first proposed network with dense connection is dense TDNN-LSTM. It has the common network skeleton with the model configuration used in \cite{Cheng17}, consisting of 7-layer time delay neural networks (TDNNs) combined with 3-layer LSTMs. The model architecture is depicted in Figure 2, where 3 TDNNs are followed by a couple of dense blocks and 1 LSTM in the final layer before the softmax layer. Each green-highlighted dense block contains 1 LSTM and 2 TDNNs with the dense connectivity pattern. The final layer in each block is to concatenate all the outputs of the neural layers inside the block.  

Table 1 shows a WER comparison between the original TDNN-LSTM \cite{Cheng17} and the proposed dense TDNN-LSTM. For this experiment, we utilized the entire training data of Fisher English Training Part 1 and 2 (LDC2004S13, LDC2005S13) and the aforementioned Switchboard corpus. The total amount of the data used for training is approximately 2,000hrs. We report the performance of the trained models on Switchboard and CallHome separately. It is noticeable in the table that there is a statistically meaningful improvement (by around 5\%, relative) on the CallHome testset by the proposed model.

\begin{figure}
    \label{fig:denseCnnBlstm}
    \begin{minipage}[b]{1.05\linewidth}
        \centering
        \includegraphics[width=8cm]{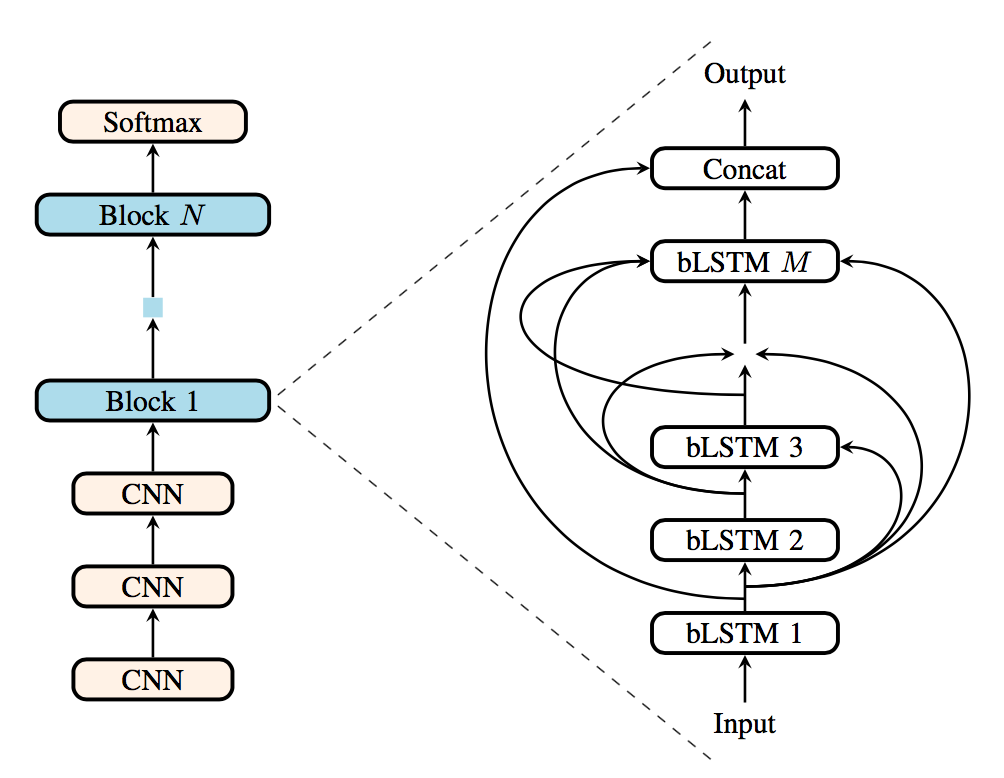}
    \end{minipage}
    \caption{Structure of a dense CNN-bLSTM acoustic model. Each dense block has 256 dimensional non-linear activation vectors.}
\end{figure}

\subsection{Dense CNN-bLSTM}
We propose another dense LSTM architecture in dense CNN-bLSTM, as shown in Figure 3. As we explore several dense CNN-bLSTMs in this paper, the figure is presented as general as possible. This architecture has 3 CNN layers followed by $N$ dense blocks (blue-highlighted), each of which contains $M$ bi-directional LSTM layers being connected densely to one another. The final layer in each block concatenates the output vectors from all the layers inside to deliver to the next block.

Table 2 presents the performances of a few dense CNN-bLSTMs with different configurations. Like the experiments for Table 1, all the dense CNN-bLSTMs in the table were trained on the 2000hr Switchboard/Fisher data. In the configurations (a), (b) and (c), the dense CNN-bLSTMs have the total 15 LSTM layers. The dense CNN-bLSTM-(a) and (b) have one transitional layer between the two blocks where 7 LSTM layers are allocated each ($N$=2, $M$=7), while the three blocks having 5 LSTM layers each ($N$=3, $M$=5) are tightly connected without a transitional layer in the configuration (c). In the dense CNN-bLSTM-(c), the cell dimension in each block gets smaller from 512 to 128 to make the entire neural network shape narrower as we go deeper. In the configuration (d), the dense CNN-bLSTM has the total 30 LSTM layers with a smaller cell dimension of 128. The dense CNN-bLSTM-(a),(b) and (c) all exceeded the performance of the dense TDNN-LSTM introduced in Section 2.1 for both of the Switchboard and CallHome testset. The performance gap between the dense CNN-bLSTMs seems to be largely contributed by the LSTM cell dimension. 

\begin{table}[t]
    \centering
    \caption{WER comparison for dense CNN-bLSTMs. $N$: number of dense blocks, $M$: number of bi-directional LSTM layers per dense block, $d$: LSTM cell dimension.}
     \vspace{3mm}
    \begin{tabular}{c|c|c}
        \hline
         \textbf{Dense CNN-bLSTM} & \textbf{SWBD} & \textbf{CH} \\
         \hline
         \hline
         \small (a) $N$=2, $M$=7, $d$=512 & \small 6.7\% & \small 12.5\%\\
         \hline
        \small (b) $N$=2, $M$=7, $d$=256 & \small 7.1\% & \small 12.5\% \\
         \hline
        \small (c) $N$=3, $M$=5, $d$=512,256,128 & \small 7.2\% & \small 12.6\% \\
         \hline
        \small (d) $N$=2, $M$=15, $d$=128 & \small 7.6\% & \small 13.4\% \\
         \hline
    \end{tabular}
    \label{tab:architectures}
\end{table}

\subsection{Acoustic Model Training} 
The Kaldi toolkit \cite{Povey11} was used to train these dense networks. Lattice-free maximum mutual information (LF-MMI) was chosen as an objective function for network training. The cross entropy objective function was also applied as an extra regularization as well as leaky HMM to avoid overfitting \cite{Povey16}. The learning rate was gradually adjusted from $10^{-3}$ to $10^{-5}$ over the course of 4 epochs. 

Prior to neural network acoustic modeling, we trained Gaussian mixture models (GMMs) within the framework of 3-state hidden Markov models (HMMs). The conventional 39-dimensional MFCC features were spliced over 9 frames and LDA was applied to project the spliced features onto a 40-dimensional subspace. Further projection was conducted through MLLT for better orthogonality. Speaker adaptive training (SAT) was applied with feature-space MLLR (fMLLR) to further refine mixture parameters in GMMs \cite{Gales97}.

The total 140K word tokens to cover the entire words contained in the training data were mapped to the PronLex pronunciation lexicon\footnote{https://catalog.ldc.upenn.edu/LDC97L20}. The phone dictionary consists of 42 non-silence phones with lexical stress markers on vowels as well as two hesitation phones, making the total phones to 44.

\section{Acoustic Model Adaptation}

Neural networks are well known to be hard for adaptation due to a huge number of parameters to be tuned, unlike statistical frameworks such as GMMs. As a result there have been alternative approaches to update only a small part of a neural network model \cite{Li10,Liao13,Himawan15,Lu16} to obtain adaptation benefits. In this paper, we propose a simple model adaptation scheme exploiting \textit{parameter averaging}. 

The idea is similar to how Kaldi's NNET3 acoustic model training handles the models updated across multiple GPUs throughout iterations \cite{Povey14}. Kaldi's NNET3 training strategy lets each GPU do stochastic gradient descent (SGD) separately with different randomized subsets of a training data and, after processing a fixed number of samples, averages the parameters of the models across all the jobs and re-distribute the result to each GPU. We borrow this concept of parameter averaging to average the parameters of a seed neural network model and its adapted version. 

In order to update a seed model with adaptation data before parameter averaging, we applied the same training technique in Section 2.3 with the LF-MMI objective function, but with no cross entropy objective. The learning rate was set to be gradually decreased from $10^{-5}$ to $10^{-7}$ over the course of 4 epochs.

\begin{table}[t]
    \centering
    \caption{Acoustic model adaptation results in WER. Before parameter averaging. The configuration indexes (a), (b) and (c) for the dense CNN-bLSTMs are inherited from Table 2.}
    \vspace{3mm}
    \begin{tabular}{c|c|c}
        \hline
         \textbf{Dense Model} & \textbf{SWBD} & \textbf{CH} \\
         \hline
         \hline
         \small TDNN-LSTM & \small 7.3\% $\rightarrow$ 7.7\% & \small 13.0\% $\rightarrow$ 12.2\% \\
         \hline
         \small CNN-bLSTM-(a) & \small 6.7\% $\rightarrow$7.3\% & \small 12.5\% $\rightarrow$ 12.2\% \\
         \hline
         \small CNN-bLSTM-(b) & \small 7.1\% $\rightarrow$ 7.5\% & \small 12.5\% $\rightarrow$ 12.1\% \\
         \hline
         \small CNN-bLSTM-(c) & \small 7.2\% $\rightarrow$ 7.9\% & \small 12.6\% $\rightarrow$ 12.2\% \\
         \hline
    \end{tabular}
    \label{tab:adaptation_before}
\end{table}

\begin{table}[t]
    \centering
    \caption{Acoustic model adaptation results in WER. After parameter averaging. The configuration indexes (a), (b) and (c) for the dense CNN-bLSTMs are inherited from Table 2.}
    \vspace{3mm}
    \begin{tabular}{c|c|c}
        \hline
         \textbf{Dense Model} & \textbf{SWBD} & \textbf{CH} \\
         \hline
         \hline
         \small TDNN-LSTM & \small 7.7\% $\rightarrow$ 7.2\% & \small 12.2\% $\rightarrow$ 12.1\% \\
         \hline
         \small CNN-bLSTM-(a) & \small 7.3\% $\rightarrow$ 6.9\% & \small 12.2\% $\rightarrow$ 12.0\% \\
         \hline
        \small  CNN-bLSTM-(b) & \small 7.5\% $\rightarrow$ 7.1\% & \small 12.1\% $\rightarrow$ 11.9\% \\
         \hline
         \small CNN-bLSTM-(c) & \small 7.9\% $\rightarrow$ 7.2\% & \small 12.2\% $\rightarrow$ 12.1\% \\
         \hline
    \end{tabular}
    \label{tab:adaptation_after}
\end{table}

We took an advantage of the CallHome American English Speech corpus (LDC97S42) for our experiments on acoustic model adaptation. According to the 2000 Hub5 Evaluation result report by NIST \cite{William00}, this corpus was listed as one of publicly available training materials. We only used a training portion of the corpus which contains 80 telephone conversations between native English speakers of around 13 speech hours. There is no overlap in data itself as well as speaker between this adaptation data and the CallHome portion of the NIST 2000 Hub5 English evaluation set\footnote{Unlike CallHome, it was reported in \cite{William00} that 36 out of 40 speakers in the Switchboard portion of the NIST 2000 Hub5 English evaluation set appeared in the conversations of the Switchboard corpora available for training. However, it was also reported that this would have a limited
effect in terms of enhancing performance.}, but it is expected for adapted models to perform better than before adaptation, at least against the CallHome testset.  

Tables 3 and 4 show the experimental results from the proposed model adaptation scheme, with and without parameter averaging. We tested the four dense LSTM models discussed in the previous section. Table 3 specifically presents the WERs from the updated models before parameter averaging. Although there exists a consistent improvement for the CallHome testset across the updated dense LSTM models, the performance against the Switchboard testset is all degraded. This indicates that the models updated with the adaptation data from the CallHome corpus have the parameters shifted toward CallHome-specific regions in a parameter space, but farther from a Switchboard-specific domain. The proposed parameter averaging method is shown in Table 4 to balance out the biases in the updated models. It seems to pull the Switchboard WERs back to the range before the model adaptation while preserving the benefit for the CallHome side. For the dense TDNN-LSTM and the dense CNN-bLSTM-(a), the slight changes in the SWBD WER are observed (7.3\% $\rightarrow$ 7.2\% \& 6.7\% $\rightarrow$ 6.9\%) from the proposed model adaptation scheme. The overall improvement for the CallHome testset across the updated models is approximately 5\% (relative).    

\section{System Descriptions}

\subsection{Other Acoustic Models}
To achieve the state-of-the-art performance on the NIST 2000 Hub5 English evaluation set, we trained 4 more acoustic models in CNN-bLSTMs across three different phonesets in addition to the aforementioned dense LSTM models. With more systems with various configurations from acoustic and lexical perspectives, we could obtain a better performance when combining such systems.  

For phoneset diversity, we exploited two more phonesets (CMU\footnote{http://svn.code.sf.net/p/cmusphinx/code/trunk/cmudict} and MSU\footnote{http://www.isip.piconepress.com/projects/switchboard/r\\eleases/sw-ms98-dict.text}) in addition to PronLex mentioned in Section 2.3. The CMU phoneset consists of 39 phones with three lexical stress markers. The MSU phoneset has 36 phones with no stress distinctions. For these three phonesets (CMU, MSU, PronLex), we trained CNN-bLSTMs with 3 CNN layers and 7 bLSTM layers, respectively. Different trees were formed for the 3 CNN-bLSTMs during the training stage, which could provide diversity to a combined system later. We also applied a different hesitation modeling for these CNN-bLSTMs from the dense LSTM models in Section 2. We used 11 distinct hesitation phones to better distinguish some hesitation utterances, such as `uh-huh' and `um-hum', instead of 2 hesitation phones in the dense LSTM models.

On top of those 3 CNN-bLSTM models with the 11 hesitation phones across the three phonesets, we built another PronLex-based CNN-bLSTM with 2 hesitations, totaling our individual acoustic model lineup to 8 (4 CNN-bLSTMs and 4 dense LSTM models) shown in Table 5.

For all the CNN-bLSTMs, log-mel cepstra were fed into the three convolutional layers and a 3$\times$3 kernel was applied with the filter size of 32 throughout the layers. The filtered signals were then passed to the 7-layer bLSTMs with the cell dimension of 1,024 after being appended with 100-dimensional i-vectors. Each neural network layer is followed by non-linear ReLU activation. 

\begin{table*}[]
    \centering
    \caption{Experimental evaluation in WER for the 8 individual systems and their combinations. The configuration indexes (a), (b) and (c) for the dense CNN-bLSTMs are inherited from Table 2. $d$: LSTM cell dimension.}
    \vspace{3mm}
    \begin{tabular}{c|c|c|c|c|c|c|c}
        \hline
    \multirow{2}{*}{\textbf{Acoustic Model}} & \multirow{2}{*}{\textbf{$d$}} & \multirow{2}{*}{\textbf{Phoneset}} & \multirow{2}{*}{\textbf{HES}} & \multicolumn{2}{c}{\textbf{SWBD}} & \multicolumn{2}{|c}{\textbf{CH}} \\
         \cline{5-8}
         & & & & \textbf{$N$-gram} & \textbf{RNN} & \textbf{$N$-gram} & \textbf{RNN} \\
         \hline
         \hline
         \small CNN-bLSTM & \small 1,024 & \small CMU & \small 11 & \small 6.8\% & \small 5.9\% & \small 11.5\% & \small 10.7\% \\
         \hline
        \small CNN-bLSTM & \small 1,024 & \small MSU & \small 11 & \small 6.8\% & \small 5.9\% & \small 11.3\% & \small \textbf{10.5\%} \\
         \hline
         \small CNN-bLSTM & \small 1,024 & \small PronLex & \small 11 & \small 6.7\% & \small 5.8\% & \small 11.6\% & \small 10.7\% \\
         \hline
         \small CNN-bLSTM & \small 1,024 & \small PronLex & \small 2 & \small 6.4\% & \small \textbf{5.6\%} & \small 11.4\% & \small 10.7\% \\
         \hline
         \small Dense CNN-bLSTM-(a) & \small 512 & \small PronLex & \small 2 & \small 6.9\% & \small 6.0\% & \small 12.0\% & \small 11.5\% \\
         \hline
         \small Dense CNN-bLSTM-(b) & \small 256 & \small PronLex & \small 2 & \small 7.1\% & \small 6.1\% & \small 11.9\% & \small 11.1\% \\
         \hline
         \small Dense CNN-bLSTM-(c) & \small 512,256,128 & \small PronLex & \small 2 & \small 7.2\% & \small 6.1\% & \small 12.1\% & \small 11.2\% \\
         \hline
         \small Dense TDNN-LSTM & \small 1,024 & \small PronLex & \small 2 & \small 7.2\% & \small 6.1\% & \small 12.1\% & \small 11.0\% \\
        \hline
         \hline
          \small 4 CNN-bLSTMs Combined & - & - & - & - & \small 5.2\% & - & \small 9.5\% \\
         \hline
          \small 4 Dense LSTMs Combined & - & - & - & - & \small 5.1\% & - & \small 9.6\% \\
         \hline
         \small System Combination & - & - & - & - & \textbf{5.0\%} & - & \textbf{9.1\%} \\
         \hline
    \end{tabular}
    \label{tab:final}
\end{table*}

\subsection{Language Models}
The 4-gram language model (LM) was trained with the open-source library of SRILM \cite{Stolcke02} on a combination of publicly available data, including Fisher English Training Part 1 and 2 (LDC2004T19, LDC2005T19), Switchboard-1 Release 2 (LDC97S62), CallHome American English Speech (LDC97T14), Switchboard Cellular Part 1 (LDC2001T14), TED-LIUM \cite{rousseau2012ted}, British Academic Spoken English (BASE) \cite{thompson2001research}, Michigan Corpus of American Spoken English (MICASE) \cite{simpson2003micase} and English Gigaword (LDC2003T05). We used this LM for the 2nd-pass LM rescoring. For the 1st pass decoding, we pruned the trained 4-gram LM with the pruning thresholds of 1.0e-8, 1.0e-7, and 1.0e-6 for bigrams, trigrams, and 4-grams, respectively. 

The RNN LM built with the CUED-RNNLM tookit \cite{chen2016cued} was trained on a subset of the aforementioned text data, consisting of only Fisher, Switchboard and CallHome with 2M sentences and 24M word tokens. We used variance regularization \cite{shi2014variance} as the optimization criterion of the objective function for the RNN LM with 1,000 nodes in each of two hidden layers. We trained two separate RNN LMs as we used the two different hesitation modeling approaches in 11 hesitations versus 2 hesitations, which resulted in differently normalized transcripts for model training. These RNN LMs across the different hesitation modeling approaches are also expected to provide a different level of diversity when combining the systems.

\subsection{System Combination}
In order to combine the systems, we applied a lattice combination that conducts a union of lattices from individual systems and searches the best path from the extended lattices using minimum Bayes risk decoding \cite{xu2011minimum}. Due to the different mix ups across the systems in terms of hesitation modeling and phoneset, we had to relabel the word list of each individual system before the lattice combination. The combination weights were found through a hyper-parameter optimization algorithm, called a tree-structured Parzen estimator \cite{Bergstra11}, using a held-out development set. 

\section{Experimental Results}

We evaluated the performance of the total 8 individual systems in 4 CNN-bLSTMs, 3 dense CNN-bLSTMs and 1 dense TDNN-LSTM across the three different phonesets against the Switchboard and CallHome testset of the NIST 2000 Hub5 English evaluation set. The performance in terms of WER, before and after RNN LM rescoring, is shown in Table 5.

Among the acoustic models, the PronLex-based CNN-bLSTM with 2 hesitation phones outperformed the other models, marking 5.6\% for Switchboard, while the MSU-based CNN-bLSTM with 11 hesitation phones reached the lowest 10.5\% for CallHome. These two numbers are the best reported WERs achieved by any single system so far in the industry. The CNN-bLSTM models obtained WERs 0.1\%-0.5\% (absolute) for Switchboard and 0.4\%-1.0\% (absolute) for CallHome lower than the dense acoustic models. Having 2 hesitations when using the PronLex phoneset appears to be a better choice to improve the robustness of the CNN-bLSTM model, but we didn't have the same pattern for the other phonesets in CMU and MSU (WERs not reported in the table). We observe that RNN LM rescoring provides consistent improvement in all the cases with absolute improvements between 0.5\% and 1.1\%, and a maximum reduction in WER of up to 8\% relative in the case of the dense TDNN-LSTM model on Switchboard (from 7.2\% to 6.1\%).

As briefly mentioned in Section 2.2, the cell dimension of LSTMs turns out to be a dominating factor to decide how the models perform. It is noticeable that any CNN-bLSTM model with the cell dimension of 1,024 has lower WER than any dense model with lesser cell dimension. Even the dense TDNN-LSTM model with uni-directional LSTMs offers lower WERs than some of the dense CNN-bLSTMs with bi-directional LSTMs (although the gaps are not huge), which can also be explained by the larger cell dimension of 1,024 in the dense TDNN-LSTM as compared to a maximum of 512 in the dense CNN-bLSTMs.

\begin{table*}[]
    \centering
    \caption{Experimental evaluation in WER for the 5 individual systems and their combinations. $d$ is the dimension of LSTM cells.}
    \vspace{-3mm}
    \begin{tabular}{c|c|c|c|c|c|c|c|c|c}
        \hline
    \multirow{2}{*}{\textbf{Acoustic Model}} & \multirow{2}{*}{\textbf{$d$}} & \multicolumn{2}{c}{\textbf{SWBD}} & \multicolumn{2}{|c}{\textbf{CH}} &
    \multicolumn{2}{|c}{\textbf{RT-02}} & \multicolumn{2}{|c}{\textbf{RT-03}} \\
         \cline{3-10}
         & & \textbf{$N$-gram} & \textbf{RNN} & \textbf{$N$-gram} & \textbf{RNN} & \textbf{$N$-gram} & \textbf{RNN} &
         \textbf{$N$-gram} & \textbf{RNN} \\
         \hline
         \hline
         \small CNN-bLSTM (CMU) & \small 1,024 & \small 6.8\% & \small 5.9\% & \small 11.5\% & \small 10.7\% & \small 10.4\%  & \small 9.2\% & \small 9.9\% & \small 9.0\% \\
         \hline
        \small CNN-bLSTM (MSU) & \small 1,024 & \small 6.8\% & \small 5.9\% & \small 11.3\% & \small 10.5\% & \small 10.0\% & \small 9.1\% & \small 9.7\% & \small 8.9\% \\
         \hline
         \small CNN-bLSTM (PronLex) & \small 1,024 & \small 6.4\% & \small 5.6\% & \small 11.4\% & \small 10.7\% & \small 9.9\% & \small 9.0\% & \small 9.8\% & \small 9.1\% \\
         \hline
         \small Dense CNN-bLSTM-(b) & \small 256 & \small 7.1\% & \small 6.1\% & \small 11.9\% & \small 11.1\% & \small 10.5\% & \small 9.6\% & \small 10.3\% & \small 9.5\% \\
         \hline
          \small Dense TDNN-LSTM & \small 1,024 & \small 7.2\% & \small 6.1\% & \small 12.1\% & \small 11.0\% & \small 10.7\% & \small 9.5\% & \small 10.5\% & \small 9.5\% \\
        \hline
         \hline
          \small 3 CNN-bLSTMs Combined & - & - & 5.1\% & - & 9.7\% & - & 8.2\% & - & 8.5\% \\
         \hline
          \small 5 System Combination & - & - & \textbf{5.0\%} & - & \textbf{9.1\%} & - & \textbf{8.1\%} & - & \textbf{8.0\%} \\
         \hline
    \end{tabular}
    \label{tab:final}
\end{table*}

The proposed dense LSTMs significantly contributed to system combination. By comparing the WERs of the 4 CNN-bLSTMs combined and that of the 8 systems including the 4 dense networks, the improvements resulted from the dense networks are shown to be approximately 5\% across the testsets. In contrast of dense and non-dense networks in a system combination perspective, the proposed dense models achieved similar performances for Switchboard and CallHome with the combined CNN-bLSTMs (5.1\% vs 5.2\% for Switchboard \& 9.6\% vs 9.5\% for CallHome). 

We achieved the same results from the 8 system combination (5.0\% for Switchboard and 9.1\% for CallHome, both of which are the best performances reported thus far) when we combined the 3 CNN-bLSTMs with CMU, MSU and PronLex (with 2 hesitation phones), dense CNN-bLSTM-(b) and dense TDNN-LSTM. The results are summarized in Table 6 with the additional evaluation results from the two data sets in RT-02 and RT-03\footnote{RT-02 and RT-03 are another evaluation sets provided by LDC from the past Rich Transcription evaluations with telephone conversations, publicly available in LDC2004S11 and LDC2007S10, respectively. They are exclusive with the NIST 2000 Hub5 English evaluation set, but some portions come from the Switchboard data collection projects.}, expecting them to offer a diverse view of our systems. 
	
\section{Experimental Results in Non-Telephony Environments}
To see how well our approach can be generalized to non-telephony environments, we have trained separate systems using the TED-LIUM release 2 \cite{ROUSSEAU12,ROUSSEAU14} and LibriSpeech \cite{panayotov2015librispeech} data. The TED-LIUM release 2 corpus consists of 212 hours of training data, a development set (1.6 hours) and a test set (2.6 hours). LibriSpeech is a 960-hour open-sourced corpus derived from read audio books, which are manually segmented and transcribed. There are 2 development sets: dev-clean (5.5 hours) and dev-other (5.1 hours), and two evaluation sets: test-clean (5.5 hours) and test-other (5.4 hours). We trained the total 8 systems across two different trees (4 systems each for a 2-state and 3-state tree) for each corpus using the training portion to evaluate against both of the dev and test data. The acoustic models trained are TDNN, TDNN-LSTM, dense TDNN-LSTM and CNN-bLSTM. The PronLex phoneset and the 2 hesitation modeling approach were exploited for all the models. The TDNN architecture has the same number of TDNN layers in the trained TDNN-LSTM models, but with no LSTM layers or dense connection. The CNN-bLSTM architecture is the same with the best performing CNN-bLSTM model in Tables 5 and 6 for the Switchboard and CallHome experiments. For the LM training (both $N$-grams and RNN-LMs), we used all the texts available in the same training portion used for the acoustic model training.

\begin{table*}[]
    \centering
    \caption{Experimental evaluation in WER for the 8 individual systems trained on TED-LIUM release 2 and their combinations. The numbers in the parentheses indicate the 2-state or 3-state trees used in the neural network acoustic model training.}
    \vspace{-3mm}
    \begin{tabular}{c|c|c|c|c}
        \hline
        \multirow{2}{*}{\textbf{Acoustic Model}} &\multicolumn{2}{c|}{\textbf{Dev}}  & \multicolumn{2}{c}{\textbf{Test}} \\
        \cline{2-5}
        & \textit{N}\textbf{-gram} & \textbf{RNN} & \textit{N}\textbf{-gram} & \textbf{RNN} \\
        \hline
        \hline
        TDNN (2)                  & 8.2\% & 7.8\% & 8.5\% & 7.8\% \\
        TDNN-LSTM (2)             & 7.8\% & 7.6\% & 8.5\% & 8.1\% \\
        CNN-bLSTM (2)            & 8.0\% & 7.7\% & 8.3\% & 7.9\% \\
        Dense TDNN-LSTM (2)       & 7.9\% & 7.5\% & 8.2\% & \textbf{7.6\%} \\
        TDNN (3)            & 8.5\% & 8.1\% & 9.0\% & 8.3\% \\
        TDNN-LSTM (3)       & 7.8\% & 7.5\% & 8.3\% & 7.9\% \\
        CNN-bLSTM (3)       & 7.8\% & 7.4\% & 8.0\% & \textbf{7.6\%} \\
        Dense TDNN-LSTM (3) & 7.6\% & \textbf{7.1\%} & 8.1\% & 7.7\% \\
        \hline
        \hline
        System combination    & - & \textbf{6.2\%} & - & \textbf{6.5\%} \\
        \hline
    \end{tabular}
    \label{tab:tedlium}
\end{table*}

\begin{table*}[]
    \centering
    \caption{Experimental evaluation in WER for the 8 individual systems trained on LibriSpeech and their combinations. The numbers in the parentheses indicate the 2-state or 3-state trees used in the neural network acoustic model training.}
    \vspace{-3mm}
    \begin{tabular}{c|c|c|c|c|c|c|c|c}
        \hline
        \multirow{3}{*}{\textbf{Acoustic Model}} &\multicolumn{4}{c|}{\textbf{Dev}}  & \multicolumn{4}{c}{\textbf{Test}} \\
        \cline{2-9}
        & \multicolumn{2}{c|}{\textbf{Clean}} & \multicolumn{2}{c|}{\textbf{Other}} &
        \multicolumn{2}{c|}{\textbf{Clean}} &
        \multicolumn{2}{c}{\textbf{Other}} \\
        \cline{2-9}
        & \textit{N}\textbf{-gram} & \textbf{RNN} & \textit{N}\textbf{-gram} & \textbf{RNN} & \textit{N}\textbf{-gram} & \textbf{RNN} & \textit{N}\textbf{-gram} & \textbf{RNN} \\
        \hline
        \hline
        TDNN (2)                  & 3.88\% & 3.51\% & 10.09\% & 9.55\% & 4.31\% & 4.07\% & 10.61\% & 10.12\% \\
        TDNN-LSTM (2)             & 3.46\% & 3.17\% & 9.90\%  & 9.68\% & 3.80\% & 3.63\% & 9.98\%  & 9.46\% \\
        CNN-bLSTM (2)            & 3.21\% & 3.08\% & 9.24\%  & 8.92\% & 3.78\% & 3.67\% & 9.54\%  & 9.21\% \\
        Dense TDNN-LSTM (2)       & 3.26\% & \textbf{3.02\%} & 9.91\%  & 8.78\% & 3.74\% & 3.56\% & 9.24\%  & 8.72\% \\
        TDNN (3)            & 3.96\% & 3.62\% & 10.29\% & 9.64\% & 4.34\% & 4.08\% & 10.67\% & 10.09\% \\
        TDNN-LSTM (3)      & 3.43\% & 3.19\% & 9.20\%  & 8.59\% & 3.83\% & 3.63\% & 9.33\%  & 8.90\% \\
        CNN-bLSTM (3)       & 3.35\% & 3.12\% & 8.78\%  & \textbf{8.28\%} & 3.63\% & \textbf{3.51\%} & 8.94\%  &\textbf{ 8.58\%} \\
        Dense TDNN-LSTM (3) & 3.39\% & 3.06\% & 9.10\%  & 8.57\% & 3.95\% & 3.65\% & 9.40\%  & 8.79\% \\
        \hline
        \hline
        System combination    & - & \textbf{2.68\%} & - & \textbf{7.56\%} & - & \textbf{3.19\%} & & \textbf{7.64\%} \\
        \hline
    \end{tabular}
    \label{tab:librispeech}
\end{table*}

Tables 7 and 8 show the evaluation results on the two non-telephony corpora. From the tables, it is hard to tell how many states in trees would be desirable when training the neural network acoustic models, but using the different trees across the models is likely to provide diversity when the systems are combined as mentioned in Sections 4 and 5. We observe that the dense TDNN-LSTMs provide the best results for TED-LIUM in both dev and test while presenting the lowest WER against dev-clean in LibriSpeech. For the other cases, the CNN-bLSTMs dominated the other models with the marginal gaps. The combined results from the total 8 systems against the corresponding test portions for both TED-LIUM and LibriSpeech are the best WERs reported thus far for the corpora. 



\section{Conclusions} 

In this paper we have proposed several densely connected LSTM architectures, bringing the dense connectivity that was successful for CNNs in image classification tasks to the LSTM framework for conversational speech recognition. This allowed LSTMs to have more direct connections between layers such that layer-wise vanishing gradient effect can be further alleviated even as more layers are stacked in a deep neural network model. Also we have introduced parameter averaging for acoustic model adaptation that averages the parameters of a seed neural network acoustic model and its adapted one, in order to balance out between domain adaptation and generalization. We have also shown a generalized performance improvement using the dense architectures for the non-telephony data sets like TED-LIUM and LibriSpeech with no corpus-specific tuning for the systems. 

We note that in a comparison of the reported numbers of 5.1\% and 9.9\% from \cite{Kurata17} our combined system for the telephony tasks has made a significant improvement on the CallHome portion of the NIST 2000 Hub5 English evaluation set, mainly due to the acoustic model adaptation using the CallHome training data of approximately 13 hours of speech. This shows that domain specific data which has similar acoustics and lexical information would have direct impact on performance improvement. 
As discussed in \cite{Han17}, even the best conversational speech recognition system could suffer from higher error rates when it is tested against real-world data with a number of unseen dynamics in data characteristics. To have systems more robust to unseen testing conditions, given limited resources of audio data and the corresponding reference transcripts, unsupervised learning that can continuously improve the recognition coverage of a given speech recognition system would be required. In addition, various testing materials beyond the NIST 2000 Hub5 English evaluation set or RTs would be able to provide deeper insights on how systems could be generalized against real-world data.

\bibliographystyle{IEEEtran/bibtex/IEEEtran}
\bibliography{ref}


\end{document}